%% file: acl_main.tex
\documentclass[11pt]{article}

\usepackage[preprint]{acl}

\usepackage{times}
\usepackage{latexsym}

\usepackage[T1]{fontenc}
\usepackage[utf8]{inputenc}

\usepackage{microtype}

\usepackage{inconsolata}

\usepackage{graphicx}

\usepackage{algorithm}
\usepackage{amsmath}
\usepackage{amsfonts} 
\usepackage{booktabs}
\usepackage[table]{xcolor}
\usepackage{xspace}
\usepackage{algpseudocode}

\newcommand{\ourmethod}{DHPO\xspace}

\definecolor{RowGray}{HTML}{F2F3F5}

\definecolor{RowBlue}{HTML}{E6F4FF}
\newcommand{\deporow}{\rowcolor{RowBlue}}

\title{Orchestrating Tokens and Sequences: Dynamic Hybrid Policy Optimization for RLVR}

\author{
 \textbf{Zijun Min\textsuperscript{1}}\thanks{Equal contribution.},
 \textbf{Bingshuai Liu\textsuperscript{1}}\footnotemark[1],
 \textbf{Ante Wang\textsuperscript{1,3}},
\\
 \textbf{Long Zhang\textsuperscript{2}},
 \textbf{Anxiang Zeng\textsuperscript{2}},
 \textbf{Haibo Zhang\textsuperscript{2}}\thanks{Corresponding author.},
 \textbf{Jinsong Su\textsuperscript{1}}\footnotemark[2]
\\
 \textsuperscript{1}School of Informatics, Xiamen University
 \textsuperscript{2}LLM Team, Shopee Pte. Ltd.
\\
 \textsuperscript{3}Institute for AI Industry Research (AIR), Tsinghua University
\\
\texttt{\{minzijun,bsliu,wangante\}@stu.xmu.edu.cn, long.zhangzl@shopee.com,}
\\
\texttt{tsengrex@sea.com, peter.wu@shopee.com, jssu@xmu.edu.cn}
}

\begin{document}
\maketitle

\input{sections/0.abstract}
\input{sections/1.introduction}

\input{sections/2.preliminary}
\input{sections/3.method}
\input{sections/4.experiments}
\input{sections/5.related_work}
\input{sections/6.conclusion}
\input{sections/7.limitations}

% \section*{Acknowledgments}

% Custom bibliography entries only
\bibliography{references}

\appendix
\input{sections/8.appendix}

\end{document}

%% file: sections/0.abstract.tex
\begin{abstract}

Reinforcement Learning with Verifiable Rewards (RLVR) offers a promising framework for optimizing large language models in reasoning tasks. 
However, existing RLVR algorithms focus on different granularities, and each has complementary strengths and limitations. 
Group Relative Policy Optimization (GRPO) updates the policy with token-level importance ratios, which preserves fine-grained credit assignment but often suffers from high variance and instability. 
In contrast, Group Sequence Policy Optimization (GSPO) applies single sequence-level importance ratios across all tokens in a response that better matches sequence-level rewards, but sacrifices token-wise credit assignment.
In this paper, we propose \textbf{D}ynamic \textbf{H}ybrid \textbf{P}olicy \textbf{O}ptimization (\textbf{\ourmethod}) to bridge GRPO and GSPO within a single clipped surrogate objective. 
\ourmethod combines token-level and sequence-level importance ratios using weighting mechanisms. 
We explore two variants of the mixing mechanism, including an \emph{averaged} mixing and an \emph{entropy-guided} mixing.
To further stabilize training, we employ a \emph{branch-specific clipping} strategy that constrains token-level and sequence-level ratios within separate trust regions before mixing, preventing outliers in either branch from dominating the update.
Across seven challenging mathematical reasoning benchmarks, experiments on both dense and MoE models from the Qwen3 series show that \ourmethod consistently outperforms GRPO and GSPO.
We will release our code upon acceptance of this paper.

\end{abstract}

%% file: sections/1.introduction.tex
\section{Introduction}

Reinforcement Learning with Verifiable Rewards (RLVR) has emerged as a central paradigm for optimizing large language models (LLMs), particularly in verifiable reasoning tasks such as mathematics and programming, where solutions can be automatically checked by rule-based verifiers.
Despite its promise, achieving stable policy optimization in RLVR remains a significant challenge.
Recent methods like Group Relative Policy Optimization (GRPO)~\cite{shao2024grpo} and Group Sequence Policy Optimization (GSPO)~\cite{zheng2025gspo} have demonstrated scalability, yet exhibit inherent limitations that affect robustness and generalization.

GRPO applies token-level importance ratios, which can be misaligned with RLVR rewards that are typically defined at the sequence level~\cite{zheng2025gspo,tan2025gtpo}. 
As training progresses, token-level importance ratios tend to exhibit high variance, making updates unstable and sensitive to outliers~\cite{zhao2025gmpo}. 
Although GRPO employs clipping to control this variance, overly tight clipping can suppress necessary exploration and cause the policy to collapse into repetitive low-diversity outputs too early.
GSPO addresses this mismatch by defining importance ratios at the sequence level, using a geometric mean over token likelihoods to stabilize optimization~\cite{zheng2025gspo}. 
However, this uniform assignment means all tokens within a sequence share the same importance ratio and advantage, which obscures fine-grained, token-level credit assignment.
Overall, GRPO provides fine-grained token-level updates but often suffers from high variance, whereas GSPO offers more stable sequence-level updates but can be overly coarse.
This analysis indicates that, within the RLVR framework, purely token-level or purely sequence-level optimization alone is inadequate for complex reasoning tasks.

To address this issue, we propose \textbf{D}ynamic \textbf{H}ybrid \textbf{P}olicy \textbf{O}ptimization (\textbf{\ourmethod}), a unified approach that integrates both perspectives within a single clipped surrogate objective.
The core idea is to replace a single-level importance ratio with a hybrid mixing of token-level and sequence-level ratios. 
We explore two variants of the mixing mechanism, including an \emph{averaged} mixing and an \emph{entropy-guided} mixing.
The token entropy under the current policy provides a lightweight uncertainty signal. 
When uncertainty is high, \ourmethod assigns greater weight to token-level ratios to preserve fine-grained local information; as confidence grows, it shifts emphasis toward sequence-level ratios to promote globally consistent updates.
To further stabilize optimization, we introduce a \emph{branch-specific clipping} strategy, which constrains the token-level and sequence-level ratios within separate trust regions before combining them. 
This prevents outlier behavior in either branch from dominating the combined update.
Together, hybrid weighting and branch-specific clipping yield trajectory-aware updates that retain the expressiveness of token-level learning while benefiting from the stability of sequence-level correction.

We evaluate \ourmethod\ within the SimpleRL framework~\cite{zeng2025simplerl} on a suite of mathematical reasoning benchmarks on dense and MoE models from the Qwen3 series~\cite{yang2025qwen3} with different scale.
Our method consistently outperforms both GRPO and GSPO in all settings.
Particularly, on the Qwen3-30B-A3B-Base, \ourmethod\ improves accuracy on AIME24 from 22.5\% (GRPO) to 34.4\%, AIME25 from 14.6\% to 26.5\%. On average, it exceeds GRPO by about 4.9\% and GSPO by 4.3\%.
These results highlight the value of adaptively balancing token-level and sequence-level optimization, rather than committing to either granularity.

%% file: sections/2.preliminary.tex
\section{Preliminaries}

In this section, we introduce GRPO and GSPO, both of which are the basis of our work.

\subsection{GRPO}

Group Relative Policy Optimization (GRPO)~\cite{shao2024grpo} is a reinforcement learning method by using group-based advantages with token-level importance ratios.
Given a query $q$, GRPO samples a group of $G$ responses $\{o_i\}_{i=1}^G$ from a behavior policy $\pi_{\theta_{\text{old}}}$.
Each response $o_i$ receives a scalar reward $r_i$.
For each query, GRPO constructs a group-relative advantage for each response as follows
\begin{equation}
A_i = \frac{r_i - \operatorname{mean}(r_1, r_2, \dots, r_G)}
{\operatorname{std}(r_1, r_2, \dots, r_G)},
\end{equation}
where $r_i$ denotes the sequence-level reward of response $o_i$. 
The same advantage $A_i$ is then uniformly assigned to all tokens within the corresponding response.

The policy update in GRPO applies PPO-style clipping at the token level. 
It utilizes the following token-level importance ratio \begin{equation}
r_{i,t}(\theta)=
\frac{\pi_{\theta}(o_{i,t}\mid q,o_{i,<t})}
{\pi_{\theta_{\text{old}}}(o_{i,t}\mid q,o_{i,<t})},
\end{equation}
where $o_{i,t}$ denotes the $t$-th token of the $i$-th response $o_i$, and $o_{i,<t}$ is its preceding context.
The clip operator is defined element-wise as $\text{clip}(x,a,b)=\min(\max(x,a),b)$, with asymmetric bounds
$1-\varepsilon_{\text{low}}^{\text{token}}$ and $1+\varepsilon_{\text{high}}^{\text{token}}$, where
$\varepsilon_{\text{low}}^{\text{token}},\varepsilon_{\text{high}}^{\text{token}}>0$ control the maximum allowable relative decrease and increase of the token-level ratio, respectively.
This yields the objective:
\begin{multline}
\mathcal{L}_{\text{GRPO}}(\theta)=
\mathbb E_{q\sim \mathcal P(Q),\,o_{i,t}\sim\pi_\theta(\cdot\mid q,\,o_{i,<t})}\Big[\\
\frac{1}{G}\sum_{i=1}^{G}\frac{1}{|o_i|}\sum_{t=1}^{|o_i|}
\min\big(r_{i,t}(\theta)A_i,\\
\text{clip}(r_{i,t}(\theta),\,1-\varepsilon_{\text{low}}^{\text{token}},\,1+\varepsilon_{\text{high}}^{\text{token}})\,A_i\big)
\Big].
\end{multline}

A central challenge with GRPO arises from the misalignment between the granularity of its off-policy correction and its supervision: while the correction is applied at the token level, the reward signal is provided only at the sequence level. This misalignment can cause token-level importance ratios to exhibit high variance, especially in long-horizon generation tasks. Consequently, the clipping strategy is frequently activated, which may constrain effective learning and limit the stability of policy updates over time~\cite{zhao2025gmpo,zheng2025gspo,tan2025gtpo}.

\subsection{GSPO}
Group Sequence Policy Optimization (GSPO)~\cite{zheng2025gspo} operates under the same group-based sampling and advantage construction framework as GRPO, but fundamentally shifts the unit of optimization from the token level to the sequence level.
Specifically, GSPO defines a length-normalized sequence-level importance ratio for each response:
\begin{equation}
s_i(\theta)=\Big(\frac{\pi_{\theta}(o_i\mid q)}{\pi_{\theta_{\text{old}}}(o_i\mid q)}\Big)^{\frac{1}{|o_i|}},
\end{equation}
where the exponent $\frac{1}{|o_i|}$ serves to mitigate the exponential growth of raw sequence likelihood ratios.
To retain token-level gradients while applying this trajectory-level scaling factor, GSPO constructs a token-wise adjustment term
\begin{equation}
s_{i,t}(\theta)=\text{sg}[s_i(\theta)]\cdot
\frac{\pi_{\theta}(o_{i,t}\mid q,o_{i,<t})}{\text{sg}[\pi_{\theta}(o_{i,t}\mid q,o_{i,<t})]},
\end{equation}
which preserves token-level gradients while enforcing the sequence-level scaling.
GSPO then applies PPO-style clipping to the sequence-level importance ratio:
\begin{multline}
    \mathcal{L}_{\text{GSPO}}(\theta)=
    \mathbb E_{q\sim \mathcal P(Q),\,o_{i,t}\sim\pi_\theta(\cdot\mid q,\,o_{i,<t})}\Big[\\
    \frac{1}{G}\sum_{i=1}^{G}
    \min\big(s_{i}(\theta)A_i,\\
    \text{clip}(s_{i}(\theta),\,1-\varepsilon_{\text{low}}^{\text{seq}},\,1+\varepsilon_{\text{high}}^{\text{seq}})\,A_i\big)
    \Big].
\end{multline}
By basing the importance correction on a trajectory-level statistic, GSPO better aligns the optimization signal with the sequence-level reward, which typically leads to more stable updates compared to GRPO.
However, this approach introduces a key limitation: because the same sequence-level ratio is applied to all tokens within a response, fine-grained credit assignment at the token level is obscured. This can be particularly detrimental in reasoning tasks, where only a critical subset of tokens determines the final outcome. 
Furthermore, to maintain stability under significant policy shifts, GSPO often requires conservative clipping thresholds, which may over-constrain updates and reduce learning efficiency.

%% file: sections/3.method.tex
\section{Methods}

In this section, we give a detailed description of \ourmethod. 
The core motivation behind \ourmethod is to jointly leverage two complementary sources of information: (i) fine-grained token-level signals, which are crucial for local credit assignment and enabling nuanced exploration, and (ii) coarse-grained sequence-level signals, which naturally align with sequence-level rewards and provide a more globally consistent correction to the policy distribution.

In the following, we first formulate the main objective of \ourmethod based on a hybrid importance ratio. 
We then describe two weighting strategies for combining token-level and sequence-level ratios. 
Finally, we introduce a branch-specific clipping strategy and analyze the resulting gradient formulation.

\subsection{Main Objective}

Motivated by the complementary strengths and respective limitations of GRPO and GSPO, we propose to replace their single-level importance ratio with a mixture of token-level and sequence-level ratios.
This design allows the update rule to smoothly interpolate between token-wise correction and sequence-wise stabilization in a data-dependent manner.

Formally, we optimize a PPO-style clipped surrogate objective defined as follows:
\begin{multline}
    \mathcal{L}_{\text{\ourmethod}}(\theta)=\mathbb E_{q\sim \mathcal P(Q),\,o_{i,t}\sim\pi_\theta(\cdot\mid q,\,o_{i,<t})}\Big[
    \\
    \frac{1}{G}\sum_{i=1}^{G}\frac{1}{|o_i|}\sum_{t=1}^{|o_i|}
    \min\Big(m_{i,t}(\theta)A_i,\,\tilde m_{i,t}(\theta) A_i\Big)\Big],
\end{multline}
where $G$ denotes the group size, $o_i$ is the $i$-th sampled response, and $A_i$ is its estimation of group advantage. 
Here, $m_{i,t}(\theta)$ is the \emph{mixed} importance ratio for the $t$-th token of the $i$-th response, while $\tilde m_{i,t}(\theta)$ is its clipped counterpart induced via a branch-specific clipping strategy detailed in Section~\ref{sec:clip}. 

Concretely, we define the mixed importance ratio as a convex combination of the token-level ratio and the sequence-level ratio:
\begin{equation}
    m_{i,t}(\theta)=w_{i,t}\,r_{i,t}(\theta)+(1-w_{i,t})\,s_{i,t}(\theta),
\end{equation}
where the token-level ratio $r_{i,t}(\theta)$ facilitates fine-grained, per-token credit assignment, whereas the sequence-level ratio $s_{i,t}(\theta)$ encapsulates the general change in the response probability and aligns directly with sequence-level rewards.
The mixing weight $w_{i,t}$$\in$$[0,1]$ controls the contribution of each component, enabling a continuous interpolation between a GRPO-like token-level update and a GSPO-like sequence-level update.
In this work, we try two ways to define $w_{i,t}$: \emph{Averaged Mixing} and \emph{Entropy-guided Mixing}, both described in the following Section~\ref{sec:dynamic}.

\subsection{Hybrid Weighting of Importance Ratios}
\label{sec:dynamic}

We mainly consider the following two definitions of $w_{i,t}$:

\paragraph{Averaged Mixing.}
We begin with a simple and canonical instantiation where the mixing weight is held constant across all tokens and samples.
Specifically, we set $w_{i,t}=0.5$, which corresponds to taking an arithmetic average of the token-level and sequence-level importance ratios.
This averaged mixing mechanism provides a straightforward, hyperparameter-free interpolation between GRPO-style token-level updates and GSPO-style sequence-level updates.
It yields a time-invariant and sample-invariant hybrid signal, serving as a robust and stable default configuration.
Due to its simplicity and consistency, averaged mixing establishes a strong baseline that already captures the complementary benefits of both granular token-wise correction and global sequence-level stabilization.

\paragraph{Entropy-guided Mixing.}
Building on the averaged formulation, we further consider a refined weighting mechanism that conditions the mixing weight on the local uncertainty of the policy.
For each sampled response $o_i$ and token position $t$, we compute the token-level entropy under the current policy:
\begin{equation}
    \mathcal H_{i,t}(\theta)=-\sum_{v\in\mathcal V}
    \pi_\theta(v\mid q,o_{i,<t})\\
    \log\pi_\theta(v\mid q,o_{i,<t}),
\end{equation}
where $\mathcal V$ is the vocabulary. 
This entropy measures the uncertainty of $\pi_\theta(\cdot\mid q,o_{i,<t})$: higher values indicate a more diffuse distribution over candidate tokens, whereas lower values indicate a more peaked distribution.

We then convert $\mathcal H_{i,t}(\theta)$ into a mixing coefficient $w_{i,t}$ via a squashed transformation:
\begin{equation}
    w_{i,t}=g(\text{sg}[\mathcal H_{i,t}(\theta)])\in[0,1],
\end{equation}
where $g(\cdot)$ denotes a min-max normalization to map entropy into a stable weighting range.

% The resulting mixing weight dynamically adjusts the emphasis between the two ratios.
% When the policy is highly uncertain ($\mathcal H_{i,t}(\theta)$ large), a larger $w_{i,t}$ favors the token-level ratio $r_{i,t}(\theta)$, preserving fine-grained local information that aids exploration. 
% Conversely, when the policy is confident ($\mathcal H_{i,t}(\theta)$ small), $w_{i,t}$ decreases, shifting emphasis toward the sequence-level ratio $s_{i,t}(\theta)$. 
% This promotes more stable updates that are better aligned with the global reward signal. Thus, entropy-guided mixing inherently implements an adaptive bias-variance trade-off, responding to the policy's local uncertainty.
Under this mechanism, the mixing weight places relatively more emphasis on the token-level ratio $r_{i,t}(\theta)$ when the policy exhibits higher uncertainty, thus preserving fine-grained local signals that support exploration.
As the policy becomes more confident, the weight shifts towards the sequence-level ratio $s_{i,t}(\theta)$, favoring updates that are more consistent with the global reward structure.
In this way, entropy-guided mixing offers a principled refinement over uniform averaging by modulating the balance between local credit assignment and global stabilization according to the policy's local uncertainty.

\subsection{Branch-Specific Clipping}
\label{sec:clip}

Clipping the importance ratio is a key stabilization technique in PPO-style policy optimization.
It constrains policy updates by truncating excessively large deviations between current and previous policies, thereby preventing a small number of high-leverage samples from dominating the gradient estimation. 
A key challenge when using hybrid importance ratios is that the token-level and sequence-level branches exhibit markedly different numerical behaviors.
Token-level ratios $r_{i,t}(\theta)$, while allowing for fine-grained correction, are prone to high variance and can become noisy. 
In contrast, sequence-level ratios $s_{i,t}(\theta)$ aggregate probability changes over entire responses, which can lead to extreme values due to multiplicative effects across long trajectories.
Applying a single, shared clipping range to the combined ratio $m_{i,t}(\theta)$ is therefore suboptimal: overly tight clipping may suppress useful local signal from $r_{i,t}(\theta)$, while overly loose clipping may fail to control instability arising from large deviations in $s_{i,t}(\theta)$.

To address this issue, we propose a \emph{branch-specific} clipping strategy that clips each ratio independently within its own trust region before mixing:
\begin{multline}
\label{eq:branch_clip}
    \tilde m_{i,t}(\theta)=w_{i,t}\,\cdot\,\text{clip}(r_{i,t}(\theta),\,1-\varepsilon^{\text{token}}_{\text{low}},\,1+\varepsilon^{\text{token}}_{\text{high}})
    \\
    +(1-w_{i,t})\cdot\text{clip}(s_{i,t}(\theta),\,1-\varepsilon^{\text{seq}}_{\text{low}},\,1+\varepsilon^{\text{seq}}_{\text{high}}).
\end{multline}
This formulation decouples the clipping coefficients $\varepsilon^{\text{token}}_{\text{low/high}}$ and $\varepsilon^{\text{seq}}_{\text{low/high}}$, independent control over the trust regions for local (token-level) and global (sequence-level) corrections independently.

Importantly, clipping prior to mixing preserves the intended semantics of each component: each ratio is constrained to remain within its own admissible update range, preventing an outlier value in one branch from unduly influencing the combined update. 
Branch-specific clipping strategy thus complements hybrid weighting by providing fine-grained stabilization for each component, enabling a more balanced bias–variance trade-off. 
Together, these mechanisms support stable yet expressive policy updates that can leverage token-level exploration when beneficial while maintaining the global consistency afforded by sequence-level optimization.

\subsection{Gradient Analysis}

We now analyze the gradient of the proposed hybrid importance ratio to explain how \ourmethod provides a unified and generalized perspective on existing policy optimization methods.
For clarity, we consider the gradient of the unclipped surrogate objective with respect to $\theta$, noting that the clipping operation does not affect the gradient form in the non-saturated regime.
Taking the derivative, we obtain
\begin{multline}
    \nabla_{\theta} \mathcal{L}_{\text{\ourmethod}}(\theta)
    =\nabla_{\theta} \mathbb{E}_{q\sim \mathcal P(Q),\,o_{i,t}\sim\pi_\theta(\cdot\mid q,\,o_{i,<t})}\Big[\\
    \frac{1}{G}\sum_{i=1}^{G}\frac{1}{|o_i|}\sum_{t=1}^{|o_i|}m_{i,t}(\theta)\hat A_i \Big] \\
    =\mathbb{E}_{q\sim \mathcal P(Q),\,o_{i,t}\sim\pi_\theta(\cdot\mid q,\,o_{i,<t})}\Big[\frac{1}{G}\sum_{i=1}^{G}\frac{1}{|o_i|}\sum_{t=1}^{|o_i|}\\
    m_{i,t}(\theta)\hat{A}_i \cdot \nabla_{\theta}\log \pi_\theta(o_{i,t}\mid q,o_{i,<t})\Big].
\end{multline}

\input{tables/main_exp}
This formulation reveals that \ourmethod\ unifies GRPO and GSPO within a single gradient framework.
In comparison, GRPO relies exclusively on token-level importance sampling, which often results in high-variance updates, while GSPO applies a uniform sequence-level correction that obscures token-wise credit assignment, \ourmethod\ smoothly interpolates between these two extremes, thereby achieving fine-grained token-level learning signals while retaining the stability advantages of sequence-level correction.

%% file: tables/main_exp.tex
\begin{table*}[!t]
\centering
\resizebox{\textwidth}{!}{
\begin{tabular}{lcccccccc}
\toprule
\multicolumn{1}{c}{\textbf{Algorithm}} & \textbf{\begin{tabular}[c]{@{}c@{}}AIME24 \\ (Avg@32)\end{tabular}} & \textbf{\begin{tabular}[c]{@{}c@{}}AIME25\\(Avg@32)\end{tabular}} & \textbf{\begin{tabular}[c]{@{}c@{}}AMC23\\(Avg@4)\end{tabular}} & \textbf{\begin{tabular}[c]{@{}c@{}}Olympiad\\Bench\end{tabular}} & \textbf{\begin{tabular}[c]{@{}c@{}}MATH\\500\end{tabular}} & \textbf{\begin{tabular}[c]{@{}c@{}}Minerva\\Math\end{tabular}} & \textbf{GSM8K} & \textbf{AVG} \\ 
\midrule

\rowcolor{RowGray}
\multicolumn{9}{c}{\textbf{\textit{Qwen3-1.7B-Base}}}   \\
GRPO & 9.0 & 7.0 & 41.2 & 33.3 & 71.8 & 30.5 & 85.5 & 39.7 \\
GSPO & 9.2 & 7.2 & 41.9 & 33.9 & 70.8 & 27.2 & 85.0 & 39.3 \\
GMPO & 12.9 & 8.5 & 49.4 & 37.3 & 73.4 & 30.1 & 84.5 & 42.3 \\ 
CISPO & 15.6 & \textbf{11.1} & 48.8 & 39.0 & 75.6 & \textbf{31.6} & 85.4 & 43.8 \\ 
\deporow
\ourmethod-A & 14.6 & 9.4 & 45.0 & 35.7 & \textbf{76.6} & \textbf{31.6} & 85.6 & 42.6\\
\deporow
\ourmethod-E & \textbf{15.9} & 9.1 & \textbf{52.5} & \textbf{39.7} & 76.4 & 30.5 & \textbf{86.1} & \textbf{44.3} \\ 
\bottomrule

\rowcolor{RowGray}
\multicolumn{9}{c}{\textbf{\textit{Qwen3-4B-Base}}}   \\
% GRPO & - & - & - & - & - & - & - & -  \\
GRPO & 21.5 & 19.9 & 65.6 & 48.0 & 83.4 & 39.3 & \textbf{94.2} & 53.1  \\
GSPO & 24.6 & 19.8 & 67.5 & 49.3 & 84.6 & 36.8 & 92.8 & 53.6  \\
GMPO & \textbf{24.9} & 18.2 & 67.5 & 49.3 & 86.6 & 37.1 & 92.7 & 53.7 \\ 
CISPO & 23.4 & 20.0 & \textbf{70.6} & \textbf{52.6} & 86.2 & 37.1 & 93.3 & 54.7 \\ 
% \deporow
% \ourmethod~ & - & - & - & - & - & - & - & - \\ 
\deporow
\ourmethod-A & \textbf{24.9} & \textbf{21.2} & 70.0 & 52.3 & \textbf{87.2} & 38.2 & 94.1 & \textbf{55.4} \\ 
\deporow
\ourmethod-E & 22.3 & 20.5 & 66.2 & 51.0 & 86.8 & \textbf{39.7} & 94.0 & 54.3 \\
\bottomrule

\rowcolor{RowGray}
\multicolumn{9}{c}{\textbf{\textit{Qwen3-30B-A3B-Base}}}   \\
GRPO & 22.5 & 14.6 & 75.0 & 51.6 & 85.6 & 39.3 & 95.0 & 54.8  \\
GSPO & 25.3 & 15.4 & 74.4 & 49.6 & 85.8 & \textbf{43.8} & 93.7 & 55.4  \\
GMPO & 30.3 & 21.5 & 75.0 & \textbf{56.7} & 90.2 & 41.9 & 95.3 & 58.7 \\ 
CISPO & 17.7 & 13.8 & 66.2 & 48.3 & 84.8 & 41.5 & 94.8 & 52.4 \\ 
\deporow
\ourmethod-A & 32.4 & 24.1 & 75.6 & 54.2 & 89.2 & \textbf{43.8} & \textbf{95.5} & 59.2 \\ 
\deporow
\ourmethod-E & \textbf{34.4} & \textbf{26.5} & \textbf{76.9} & 52.3 & \textbf{92.4} & 40.8 & 94.8 & \textbf{59.7} \\
\bottomrule

\end{tabular}
}
\caption{Overall model performance across models. \ourmethod-A represents \ourmethod with averaged mixing, while \ourmethod-E represents \ourmethod with entropy-guided mixing. The results highlight the consistent improvements brought by \ourmethod. The \textbf{bold} represents the best performance among algorithms.}
\label{table:main_performance}
\vspace{-12pt}
\end{table*}

%% file: sections/4.experiments.tex
\section{Experiments}

\subsection{Setup}

All models are trained using \textsc{verl} framework~\cite{sheng2024verl} with vLLM~\cite{kwon2023vllm} as the rollout engine. 
We deploy training on a cluster of 4 nodes, each equipped with 8×NVIDIA H100 GPUs (32 GPUs in total).
For policy training, we use the SimpleRL dataset (8,192 examples)~\cite{zeng2025simplerl}. Input prompts are truncated to a maximum of 1,024 tokens, and model responses are limited to 4,096 tokens.
During rollouts, we employ a prompt batch size of 512, with each prompt generating 16 responses at a temperature of 1.0.
The actor learning rate is $1\times10^{-6}$.

We evaluate the performance of models on a suite of math reasoning benchmarks using the SimpleRL framework~\cite{zeng2025simplerl} using a decoding temperature of 1.0.
The benchmarks include AIME 2024/2025~\cite{aime24}, AMC 2023~\cite{amc23}, OlympiadBench~\cite{he2024olympiadbench}, MATH-500~\cite{hendrycks2021math500}, Minerva Math~\cite{lewkowycz2022minerva}, and GSM8K~\cite{cobbe2021gsm8k}.
For all tasks, the maximum response length is set to 16K tokens.
The results of AIME24/25 are reported as Pass@1 averaged over 32 samplings (Avg@32), while the results of AMC23 are reported as Pass@1 averaged over 4 samplings (Avg@4).
The complete hyperparameter and evaluation details are provided in Appendix~\ref{sec:hyper}.
We compare \ourmethod\ with representative RLVR algorithms, including GRPO, GSPO, GMPO, and CISPO, under the same training and evaluation settings.
Detailed descriptions of the compared RLVR algorithms and their differences are provided in Appendix~\ref{sec:baseline}.

\begin{figure*}
    \centering
    \includegraphics[width=\linewidth]{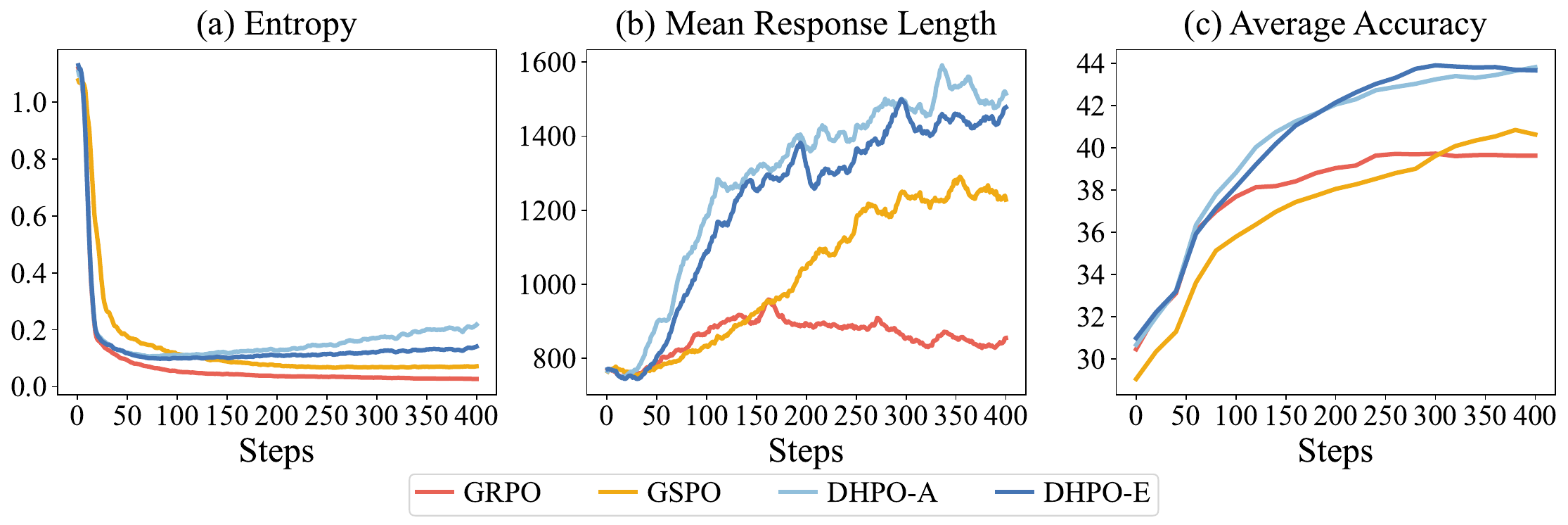}
    \caption{(a) Mean response length across training steps on Qwen3-1.7B-Base for different algorithms. GRPO collapses to shorter responses, while GSPO gradually increases response length. \ourmethod yields consistently longer and more stable responses.
    (b) Training dynamics of policy entropy on Qwen3-1.7B-Base over training steps. All methods exhibit a rapid entropy drop in the early stage. Compared with GRPO and GSPO, \ourmethod maintains consistently higher entropy in the later stage.
    (c) Average accuracy over seven benchmarks on Qwen3-1.7B-Base across training steps. GRPO and GSPO improve steadily but exhibit larger fluctuations and lower final performance. In contrast, \ourmethod achieves consistently higher accuracy. }
    \label{fig:len_entr_acc}
\end{figure*}

\subsection{Main Results}

Table \ref{table:main_performance} presents the comprehensive performance comparison across three model scales: Qwen3-1.7B-Base, Qwen3-4B-Base, and Qwen3-30B-A3B-Base. Across nearly all benchmarks and model sizes, our proposed \ourmethod variants (\ourmethod-A representing \ourmethod with averaged mixing and \ourmethod-E representing \ourmethod with entropy-guided mixing) consistently outperform the baseline methods, demonstrating the robustness and scalability of our framework~\cite{yang2025qwen3}.

On Qwen3-1.7B-Base, \ourmethod achieves notable improvements on several challenging benchmarks. 
Specifically, \ourmethod with entropy-guided mixing increases the accuracy on AIME24 from 9.0\% (GRPO) to 15.9\%, on AMC23 from 41.2\% to 52.5\%, and on OlympiadBench from 33.3\% to 39.7\%, surpassing the baseline of GRPO by approximately 4.6\% on average.
These results indicate that our entropy-guided mixing effectively balances fine-grained token-level updates with the stabilizing influence of sequence-level updates, leading to more stable convergence and better generalization even for smaller-capacity. 
A similar trend holds for the larger Qwen3-30B-A3B-Base model, particularly on challenging benchmarks such as AIME24 and AIME25, where \ourmethod with entropy-guided mixing attains 34.4\% and 26.5\%, respectively—demonstrating the method’s scalability and robustness across model sizes.

Overall, \ourmethod consistently outperforms GRPO and GSPO across diverse reasoning tasks and model scales. 
The improvements are especially evident on challenging mathematical reasoning benchmarks.
This entropy-guided hybrid mechanism is more effective than rigidly committing to a single granularity, leading to more robust and generalizable policy optimization in verifiable reasoning tasks.

\subsection{Analysis}
\label{sec:performance_analysis}

\paragraph{From Exploration to Accuracy.}
Figure~\ref{fig:len_entr_acc} reveals a unified training dynamic rather than three independent curves.
When entropy does not collapse prematurely, the policy continues to sample diverse trajectories.
This sustained diversity is crucial for supporting long-horizon reasoning, which is reflected in the increasing and stable response lengths observed during training.
In turn, longer trajectories improve performance on verifiable reasoning tasks, where multi-step derivations and intermediate computations are often necessary to reach a correct solution.
The overall trend therefore follows a clear chain: preserved entropy enables continued exploration, exploration sustains longer reasoning, and longer reasoning leads to higher averaged accuracy.

\paragraph{Stabilization with Hybrid Clipping.}
In RLVR, rewards are defined at the sequence level, yet GRPO applies token-level PPO-style reweighting with symmetric clipping (e.g. $\varepsilon_{\text{low}}$=$\varepsilon_{\text{high}}$=$0.2$).
Under such clipping, high-probability tokens are driven extremely close to $1$, whereas low-probability tokens are tightly capped and struggle to receive meaningful probability updates~\cite{yu2025dapo,yang2025dcpo}.
This asymmetric effect accelerates the concentration of the policy distribution.
As training progresses, sampled responses within each group become increasingly similar, eroding the contrast of group-based advantages and degrading the quality, as evidenced by the rapid collapse of both entropy and mean response length in Figure~\ref{fig:len_entr_acc}(a,b). 

\ourmethod\ mitigates this entropy collapse by stabilizing importance weighting while preserving exploratory capacity. 
We clip both token-level and sequence-level ratios with decoupled ranges ($[0.2,0.28]$ for both branches~\cite{yu2025dapo}). 
This allows for genuinely useful but initially unlikely actions to gain probability mass more easily, preventing the policy from prematurely over-committing to a narrow mode. 
Together, these design choices help maintain a non-deterministic policy (Figure~\ref{fig:len_entr_acc}(a)), which in turn supports longer trajectories (Figure~\ref{fig:len_entr_acc}(b)) and yields higher average accuracy (Figure~\ref{fig:len_entr_acc}(c)).

\paragraph{Entropy-Guided vs.\ Averaged Mixing}
The two hybrid variants of \ourmethod differ primarily in how they mix token-level and sequence-level importance ratios throughout training. 
The averaged mixing uses a fixed weight on token-level ratios ensuring that token-level signals continue to contribute even in later training stages, thereby maintaining stronger exploratory pressure.
In contrast, the entropy-guided mixing dynamically adjusts the mixing weight based on local policy uncertainty: it assigns greater weight to token-level ratios when uncertainty is high, and gradually shifts emphasis toward sequence-level ratios as the policy becomes more confident. 
The two variants emphasize different priorities, where the averaged mixing prioritizes sustained exploration, while the entropy-guided mixing emphasizes adaptive variance control, avoiding a return to entropy collapse.
Importantly, both variants operate within a \emph{healthy} regime where the entropy does not vanish and the response length does not regress. 
Consequently, both successfully translate sustained exploration into consistent accuracy gains across all seven benchmarks, as shown in Figure~\ref{fig:len_entr_acc}(c).

\subsection{Ablation Study}

\begin{figure}[t]
    \centering
    \includegraphics[width=\linewidth]{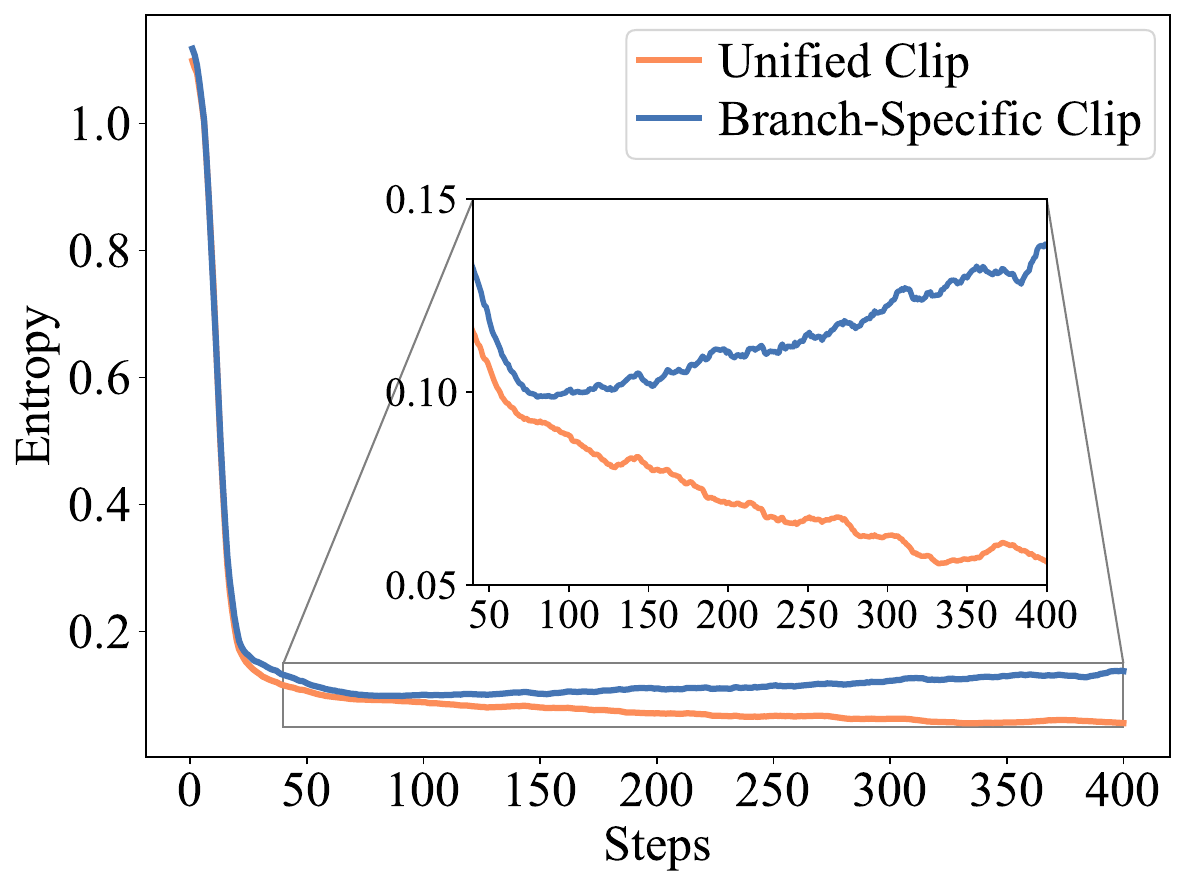}
    \caption{Training dynamics of policy entropy on Qwen3-1.7B-Base across training steps. Compared to \textbf{Unified Clip}, \textbf{Branch-Specific Clip} prevents outlier ratios from dominating the update and preserves exploration, yielding consistently higher entropy in the later stage.}
    \label{fig:entropy}
\end{figure}

To investigate the effectiveness of each component in \ourmethod,
we compare two configurations:
1) \textbf{Unified Clip}, which applies a single clipping range directly to the \emph{mixed} importance ratio; 2) \textbf{Branch-Specific Clip}, which clips token-level and sequence-level ratios in separate trust regions before mixing (Equation~\ref{eq:branch_clip}). 
The core motivation for branch-specific clipping is that it explicitly bounds each component within its own admissible range, preventing an outlier ratio in one branch from dominating the combined update and thus yielding a more stable and balanced gradient signal.

\begin{figure}[t]
    \centering
    \includegraphics[width=0.95\linewidth]{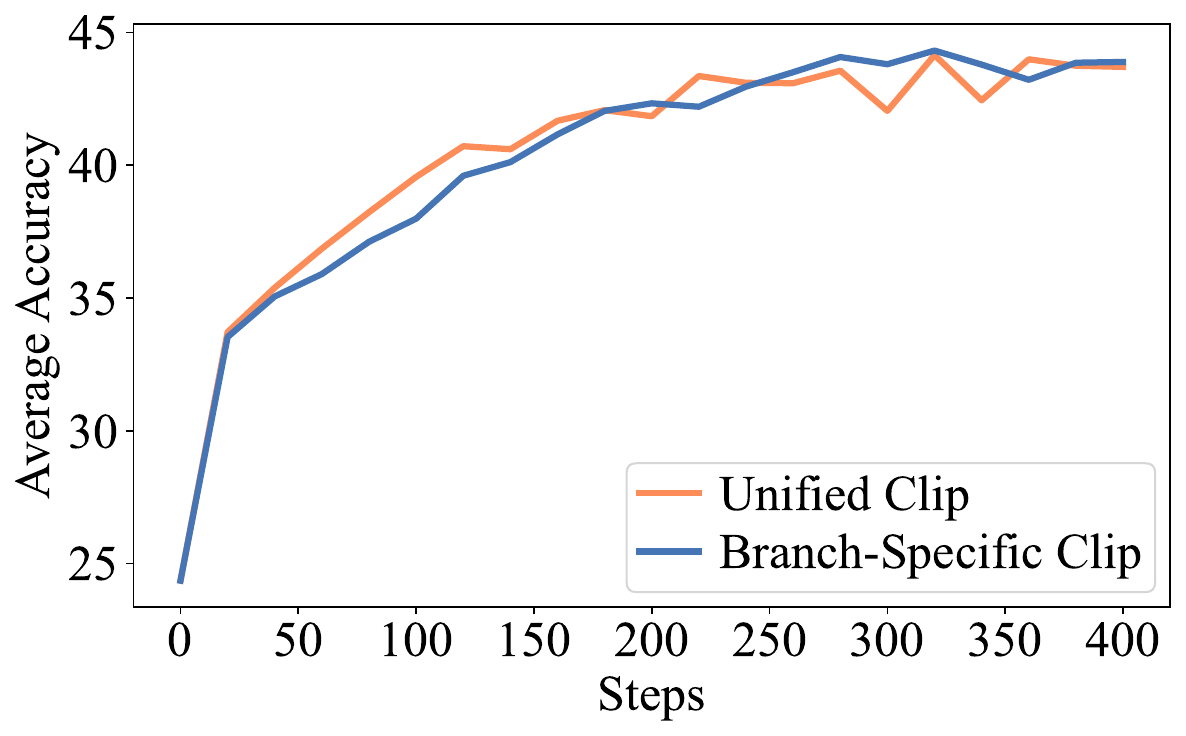}
    \caption{Averaged accuracy over seven benchmarks on Qwen3-1.7B-Base across training steps. While both methods reach similar final performance, \textbf{Branch-Specific Clip} exhibits noticeably smoother progress with smaller oscillations than \textbf{Unified Clip}, suggesting that clipping token-level and sequence-level ratios in separate trust regions reduces update variance and yields more stable optimization.}
    \label{fig:acc_avg}
\end{figure}

Empirically, the branch-specific design improves stability and preserves exploration. 
As shown in Figure~\ref{fig:entropy}, although both methods exhibit a similar initial drop in policy entropy, Branch-Specific Clip maintains consistently higher entropy in later training stages. 
This indicates a sustained exploratory capacity and a reduced tendency toward premature deterministic behavior. 
In terms of final tasks performance, Figure~\ref{fig:acc_avg} shows that both configurations achieve comparable average accuracy across seven benchmarks. 
However, Branch-Specific Clip progresses more smoothly, with noticeably smaller oscillations throughout training, suggesting reduced update variance and more stable optimization dynamics. 
Detailed training curves for each benchmark are provided in Appendix Figure~\ref{fig:acc_all_data}, which further confirms that Branch-Specific Clip broadly follows the same upward trend as Unified Clip, while exhibiting less oscillatory updates on several datasets.

%% file: sections/5.related_work.tex
\section{Related Work}

Reinforcement learning (RL) for language models is commonly built upon policy-gradient methods such as REINFORCE~\cite{williams1992reinforce} and Proximal Policy Optimization (PPO)~\cite{schulman2017ppo}, where a learned reward model or a rule-based verifier provides training signals for policy updates.
To reduce the cost and instability of PPO-style training, GRPO~\cite{shao2024grpo} removes the critic and estimates advantages through group-wise comparisons among sampled responses.
% \\
Beyond GRPO, recent RLVR methods improve stability, credit assignment, and length sensitivity.
DAPO~\cite{yu2025dapo} decouples clipping to address entropy collapse and uses dynamic sampling; it further introduces over-length penalties to better handle truncation and over-long generations.
Along the same line, several works address entropy collapse/explosion with more direct control mechanisms.
QAE~\cite{wu2025qae} replaces mean-based group advantages with a $k$-quantile estimator to reduce over-penalization of negative-advantage samples;
SIREN~\cite{jiang2025siren} applies selective entropy regularization via entropy masking to focus on a meaningful subset of tokens;
AEPO~\cite{wang2025aepo} controls entropy using temperature-adjusted regularization toward a desired entropy regime;
STEER~\cite{hao2025steer} adaptively reweights tokens based on entropy-change trends to prevent overly rapid entropy decay.
% To mitigate length inflation and balance attributes such as accuracy vs.\ conciseness, GFPO~\cite{shrivastava2025gfpo} trains on filtered rollouts that satisfy desired properties.
% GRPO-MA~\cite{wang2025grpoma} densifies feedback by generating multiple answers per shared thought process, reducing advantage variance.
% RiskPO~\cite{ren2025riskpo} adopts risk-sensitive objectives to emphasize difficult instances.
% \\
To better match sequence-level rewards, GSPO~\cite{zheng2025gspo} shifts optimization to the sequence level, while GMPO~\cite{zhao2025gmpo} uses a geometric-mean aggregation to suppress token-level outliers; FSPO~\cite{mao2025fspo} further proposes length-fair clipping for sequence objectives.
Also, GTPO~\cite{tan2025gtpo} assigns token-level entropy-weighted rewards and GRPO-S~\cite{tan2025gtpo} adopts sequence-level entropy-weighted rewards to align optimization units with reward signals.
Clipping has also been refined to control high-variance updates, including token-adaptive ranges in DCPO~\cite{yang2025dcpo}, clipping importance-sampling weights in CISPO~\cite{chen2025cispo}, and reintroducing gradients from clipped tokens in GPPO~\cite{su2025gppo}.
% \\
To the best of our knowledge, our work is the first attempt to unify token-level and sequence-level policy optimization within a single clipped surrogate objective via a hybrid importance ratio, enabling a continuous interpolation between fine-grained credit assignment and sequence-level stabilization.

%% file: sections/6.conclusion.tex
\section{Conclusion}

Reinforcement Learning with Verifiable Rewards (RLVR) has shown strong potential for improving LLMs on mathematical reasoning, but existing methods face a trade-off between high-variance token-level importance ratios and overly coarse sequence-level ones.
To leverage the complementary strengths of both perspectives, we propose \textbf{D}ynamic \textbf{H}ybrid \textbf{P}olicy \textbf{O}ptimization (\ourmethod), which unifies token-level and sequence-level importance ratios within a single clipped surrogate objective.
\ourmethod supports both a simple \emph{averaged} mixing strategy and an \emph{entropy-guided} variant that adapts the balance between the two granularities over the course of training.
To further enhance robustness, we introduce branch-specific clipping to constrain each ratio branch in its own trust region before mixing.
% To further improve robustness, we introduce branch-specific clipping, which constrains the two ratio branches in separate trust regions before mixing, preventing outliers in either branch from dominating optimization.
Across seven mathematical reasoning benchmarks and three model scales, \ourmethod consistently outperforms GRPO and GSPO, while exhibiting healthier training dynamics and smoother performance improvements.
% Across seven mathematical reasoning benchmarks and three model scales, \ourmethod consistently outperforms GRPO and GSPO, while exhibiting healthier training dynamics with sustained entropy, longer reasoning trajectories, and smoother accuracy improvements.
% Overall, our results suggest that integrating token-level and sequence-level perspectives is a practical path toward more stable and generalizable RLVR policy optimization.

% Moreover, our formulation provides a unifying view that connects token-level and sequence-level policy optimization under a single clipped objective.

%% file: sections/7.limitations.tex
\section*{Limitations}

\ourmethod\ is empirically effective, but our evaluation remains limited in scope.
Although we observe consistent gains on three Qwen3 backbones (1.7B, 4B, and 30B-A3B), our experiments are restricted to a single model family with shared pretraining objectives and tokenizer design.
We do not extend the study to other architectures or pretraining recipes, such as models with different tokenization schemes, scaling behaviors, or architectural variants, which may interact differently with hybrid importance weighting.
As a result, the generality of our findings across broader LLM ecosystems remains to be further validated.
In addition, due to computational constraints, we compare \ourmethod\ against a focused set of representative RLVR baselines rather than conducting an exhaustive evaluation over the full spectrum of RLVR algorithms and stabilization techniques.
While these baselines cover both token-level and sequence-level optimization paradigms, a more comprehensive comparison could reveal additional insights into how hybrid importance mixing interacts with alternative variance-reduction or clipping strategies.

%% file: sections/8.appendix.tex
\clearpage
\section{Detailed Gradient Analysis}
\label{sec:appendix_gradient}

In this section, we provide a detailed derivation of the policy gradient for \ourmethod.
For clarity of exposition, we omit the clipping operator and focus on the unclipped surrogate objective, as the gradient form remains unchanged within the non-saturated region.

Starting from the definition of the objective, the gradient can be written as
\begin{multline}
    \nabla_{\theta} \mathcal{L}_{\text{\ourmethod}}(\theta)
    =\nabla_{\theta} \mathbb{E}_{q\sim \mathcal P(Q),\,o_{i,t}\sim\pi_\theta(\cdot\mid q,\,o_{i,<t})}\Big[\\
    \frac{1}{G}\sum_{i=1}^{G}\frac{1}{|o_i|}\sum_{t=1}^{|o_i|}
    m_{i,t}(\theta)\hat A_i
    \Big].
\end{multline}
By exchanging the gradient and expectation under standard regularity assumptions, we obtain
\begin{multline}
    \nabla_{\theta} \mathcal{L}_{\text{\ourmethod}}(\theta)
    =\mathbb{E}_{q\sim \mathcal P(Q),\,o_{i,t}\sim\pi_\theta(\cdot\mid q,\,o_{i,<t})}\Big[\\
    \frac{1}{G}\sum_{i=1}^{G}\frac{1}{|o_i|}\sum_{t=1}^{|o_i|}
    \nabla_{\theta} m_{i,t}(\theta)\hat A_i \Big].
\end{multline}

Recall that the mixed importance ratio is defined as
$m_{i,t}(\theta)=w_{i,t}\,r_{i,t}(\theta)+(1-w_{i,t})\,s_{i,t}(\theta)$.
Since the mixing weight $w_{i,t}$ is treated as a fixed coefficient with respect to $\theta$, it does not contribute to the gradient.
Therefore, the gradient of the mixed ratio can be written as
\begin{multline}
    \nabla_{\theta} m_{i,t}(\theta)
    = m_{i,t}(\theta)\nabla_{\theta}\log m_{i,t}(\theta)\\
    = m_{i,t}(\theta)\nabla_\theta\log\big(w_{i,t}\,r_{i,t}(\theta)+(1-w_{i,t})\,s_{i,t}(\theta)\big)\\
    = m_{i,t}(\theta)\frac{
    w_{i,t}\nabla_\theta r_{i,t}(\theta)+(1-w_{i,t})\nabla_\theta s_{i,t}(\theta)
    }{
    w_{i,t}\,r_{i,t}(\theta)+(1-w_{i,t})\,s_{i,t}(\theta)
    }.
\end{multline}
Notably, both the token-level ratio $r_{i,t}(\theta)$ and the sequence-level ratio $s_{i,t}(\theta)$ depend on $\theta$ only through the same policy likelihood term $\pi_\theta(o_{i,t}\mid q,o_{i,<t})$.
As a result, their gradients share an identical score-function form, yielding
\begin{equation}
    \nabla_{\theta} m_{i,t}(\theta)
    = m_{i,t}(\theta)\nabla_{\theta}\log \pi_\theta(o_{i,t}\mid q,o_{i,<t}).
\end{equation}

Substituting this result back into the objective gradient, we finally obtain
\begin{multline}
    \nabla_{\theta} \mathcal{L}_{\text{\ourmethod}}(\theta) =\mathbb{E}_{q\sim \mathcal P(Q),\,o_{i,t}\sim\pi_\theta(\cdot\mid q,\,o_{i,<t})}\Big[\\
    \frac{1}{G}\sum_{i=1}^{G}\frac{1}{|o_i|}\sum_{t=1}^{|o_i|}
    m_{i,t}(\theta)\hat{A}_i \\
    \nabla_{\theta}\log \pi_\theta(o_{i,t}\mid q,o_{i,<t})\Big].
\end{multline}

% For notational simplicity, we denote the score function of the policy at token position $t$ as
% \begin{align}
%     \phi_\theta=\nabla_{\theta}\log\pi_{\theta}(o_{i,t}\mid q,o_{i,<t}).
% \end{align}
% Using this notation, the gradient of the surrogate objective can be rewritten:
% \begin{multline}
%     \nabla_{\theta} \mathcal{L}_{\text{\ourmethod}}(\theta)
%     =
%     \mathbb{E}_{q\sim \mathcal P(Q),\,o_{i,t}\sim\pi_\theta(\cdot\mid q,\,o_{i,<t})}\Big[\\
%     \frac{1}{G}\sum_{i=1}^{G}\frac{1}{|o_i|}\sum_{t=1}^{|o_i|}
%     m_{i,t}(\theta)\,\hat A_i \cdot \phi_\theta
%     \Big].
% \end{multline}

This expression highlights that \ourmethod\ preserves the standard policy-gradient structure, with the mixed importance ratio $m_{i,t}(\theta)$ acting as a multiplicative modulation on the score function.
Compared to GRPO, which relies solely on token-level ratios and GSPO, which assigns a uniform sequence-level weight to all tokens, our formulation interpolates between the two extremes while retaining a unified gradient form.
As a result, the proposed method enables fine-grained credit assignment at the token level without sacrificing the stability benefits of sequence-level correction.

\section{Detailed Performance}
\label{sec:detailed_performance}

To complement the averaged trends in Figure~\ref{fig:len_entr_acc}(c) and Figure~\ref{fig:acc_avg},
we report the training dynamics across benchmarks on Qwen3-1.7B-Base.
These results show how performance evolves with training steps across benchmarks, and consistently highlight the advantages of our method.

\begin{figure*}[t]
    \centering
    \includegraphics[width=\linewidth]{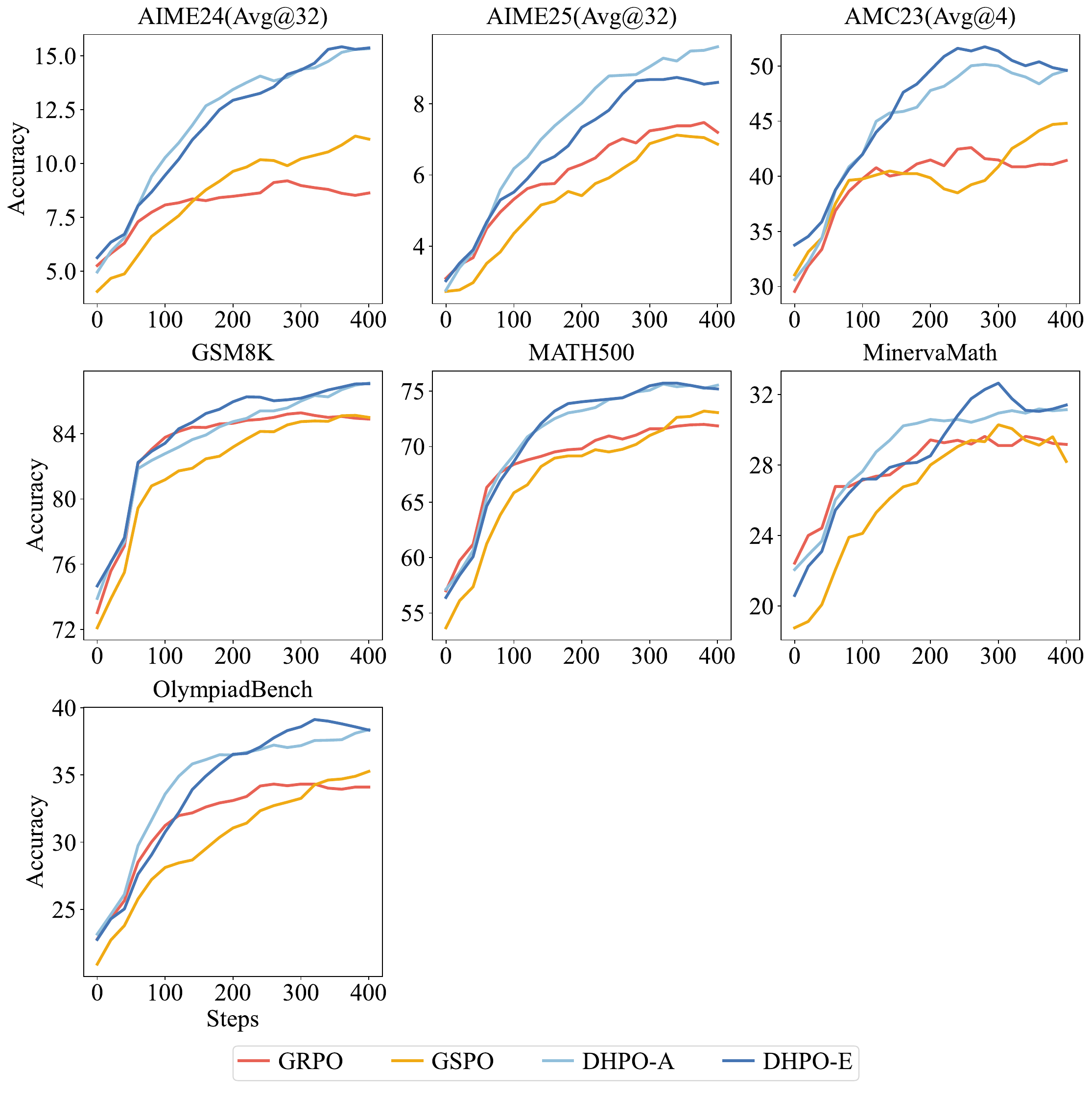}
    \caption{Training curves of accuracy over training steps on Qwen3-1.7B-Base over seven math benchmarks. \ourmethod-A represents \ourmethod with averaged mixing, while \ourmethod-E represents \ourmethod with entropy-guided mixing.}
    \label{fig:acc_all_data}
\end{figure*}

\paragraph{Training dynamics across benchmarks for main methods.}
Figure~\ref{fig:acc_all_data} presents accuracy dynamics for GRPO, GSPO, and our two variants (\ourmethod-A and \ourmethod-E) across seven math benchmarks.
Across tasks, both \ourmethod\ variants exhibit steadier improvement and reach higher plateaus.
The improvements are most evident on harder benchmarks such as AIME24/25 and OlympiadBench, where effective learning depends on sufficient exploration.
On relatively easier benchmarks such as GSM8K and MATH500, most methods converge to similarly saturated regions, but \ourmethod\ typically reaches these regions faster.
Overall, \ourmethod\ exhibits more consistent progress, benefiting from the entropy-guided mixing mechanism in Section~\ref{sec:dynamic} and the stabilization effect of branch-specific clipping in Section~\ref{sec:clip}, reflecting the exploration-to-accuracy relationship in Section~\ref{sec:performance_analysis}.

\begin{figure*}[t]
    \centering
    \includegraphics[width=\linewidth]{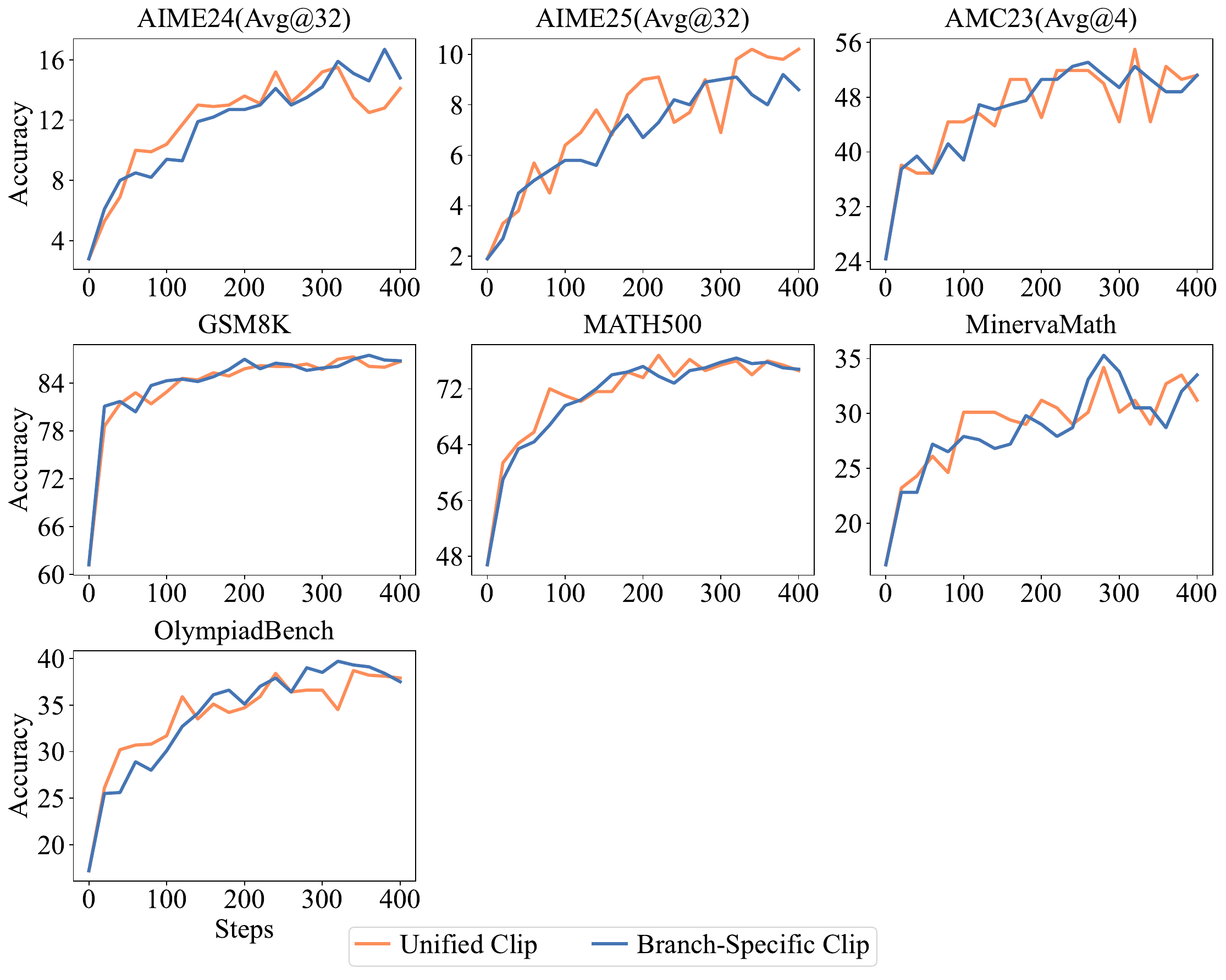}
    \caption{Training curves of accuracy over training steps on Qwen3-1.7B-Base over seven math benchmarks. \textbf{Branch-Specific Clip} generally follows the same upward trend as \textbf{Unified Clip} while exhibiting smoother, less oscillatory updates on several datasets, consistent with reduced variance from clipping token-level and sequence-level importance ratios in separate trust regions before mixing.}
    \label{fig:clip_acc_all_data}
\end{figure*}

\paragraph{Training dynamics across benchmarks for clipping ablation.}
Figure~\ref{fig:clip_acc_all_data} presents accuracy dynamics of the clipping ablation across seven math benchmarks.
Across all benchmarks, \emph{Branch-Specific Clip} largely tracks the same overall upward trend as \emph{Unified Clip}, while exhibiting smoother trajectories with smaller oscillations on several datasets (e.g., AIME and MinervaMath), where ratio outliers and long-horizon effects can amplify update variance.
This disaggregated view is consistent with the averaged behavior in Figure~\ref{fig:acc_avg}.
Clipping token-level and sequence-level ratios in separate trust regions reduces oscillations without lowering final performance.

\section{Baselines}
\label{sec:baseline}

We compare \ourmethod with four representative policy optimization baselines for RLVR. 

\begin{itemize}
    \item \textbf{GRPO}~\cite{shao2024grpo} is a PPO-style, value-free objective that uses \emph{token-level} importance ratios to reweight per-token gradients. It constructs a group-relative advantage from sequence-level rewards and assigns the same advantage to all tokens in a sampled response. Policy updates are stabilized through token-level ratio clipping.

    \item \textbf{GSPO}~\cite{zheng2025gspo} shifts the optimization unit to the sequence level to better align with sequence-level rewards. It defines a length-normalized sequence-level importance ratio and distributes it uniformly across tokens. This design typically yields more stable updates than token-level methods, but obscures fine-grained, per-token credit assignment. GSPO often employs conservative clipping thresholds to maintain stability, which may constrain update magnitudes.

    \item \textbf{GMPO}~\cite{zhao2025gmpo} modifies how token-level importance terms are aggregated within the PPO framework. Instead of using a simple arithmetic mean, it employs a geometric mean to aggregate token-level ratios. This formulation aims to down-weight the influence of outlier ratios, thereby reducing variance and improving training stability under high-variance importance sampling.

    \item \textbf{CISPO}~\cite{chen2025cispo} is a clipped importance-sampling policy optimization method that explicitly and often asymmetrically truncates importance ratios to control high-variance updates. Compared to standard PPO-style clipping, CISPO places a stronger emphasis on bounding extreme ratio values, making it a practical baseline where verifier signals are sequence-level while policy gradients are estimated from sampled trajectories.
\end{itemize}

\section{Hyperparameters and Evaluation Details}
\label{sec:hyper}

This section summarizes the key hyperparameters used in our RLVR training and evaluation.
All methods are implemented in the same \textsc{verl}~\cite{sheng2024verl} training pipeline with identical data, rollout, optimization, and evaluation settings.

\paragraph{Clipping ranges.}
We follow the recommended clipping ranges reported in the original papers of the baseline RLVR algorithms.
For \ourmethod, we use separate trust regions for the token-level and sequence-level ratios, and set 
$\varepsilon^{\text{token}}_{\text{low}}=\varepsilon^{\text{seq}}_{\text{low}}=0.2$ and
$\varepsilon^{\text{token}}_{\text{high}}=\varepsilon^{\text{seq}}_{\text{high}}=0.28$.
For GRPO, we set $\varepsilon_{\text{low}}=\varepsilon_{\text{high}}=0.2$.
For GSPO, we set $\varepsilon_{\text{low}}=3\times10^{-4}$ and $\varepsilon_{\text{high}}=4\times10^{-4}$.
For GMPO, we set $\varepsilon_{\text{low}}=\varepsilon_{\text{high}}=0.4$.
For CISPO, we set $\varepsilon_{\text{low}}=10$ and $\varepsilon_{\text{high}}=0.2$.
Apart from these clipping ranges, all remaining hyperparameters are shared across methods to isolate the effect of the surrogate objective.

\paragraph{Rollout and sequence lengths.}
We truncate input prompts to at most $1{,}024$ tokens and cap training responses at $4{,}096$ tokens.
During rollouts, each update uses a prompt batch size of $512$, and we sample $16$ responses per prompt with temperature $1.0$.
We use nucleus sampling with $\text{top-}p=1.0$ and do not apply a $\text{top-}k$ cutoff.
For evaluation, we increase the maximum response length to $16\mathrm{K}$ tokens for all benchmarks.

\paragraph{Optimization and PPO updates.}
The actor optimizer uses a learning rate of $1\times10^{-6}$ with $10$ warmup steps and weight decay $0.1$.
For Qwen3-1.7B-Base and Qwen3-4B-Base, we perform PPO-style updates with mini-batch size $256$ and micro-batch size $16$ per GPU.
And for Qwen3-30B-A3B-Base, we perform PPO-style updates with mini-batch size $32$ and micro-batch size $32$ per GPU.
We enable dynamic batching for log-probability computation and loss aggregation to better utilize GPU memory under variable-length sequences.

\paragraph{Regularization and system settings.}
We do not add an explicit KL loss term during training and also do not incorporate KL into the reward.
We enable gradient checkpointing and remove padding in the model forward pass.
We keep parameter offloading and optimizer offloading disabled in FSDP to avoid additional communication overhead in our setting.
All runs follow the same logging, validation, and checkpoint schedule, and we report results under the same evaluation protocol described in the main text.